\documentclass[letterpaper, 10 pt, conference]{ieeeconf}

\usepackage[hidelinks]{hyperref}
\usepackage{amssymb}
\usepackage{ifthen}
\usepackage{tikz}
\usepackage{multirow}
\usepackage[table]{xcolor}
\usepackage{subcaption}
\usepackage[strict]{changepage}
\usepackage{framed}

\newboolean{showcomments}
\setboolean{showcomments}{true}
\ifthenelse{\boolean{showcomments}}
 { \newcommand{\mynote}[2]{
      \fbox{\bfseries\sffamily\scriptsize#1}
        {\small$\blacktriangleright$\textsf{\emph{#2}}$\blacktriangleleft$}}}
        { \newcommand{\mynote}[2]{}}

\definecolor{formalshade}{rgb}{0.85,1,0.85}
\definecolor{darkblue}{rgb}{0.0,0.6,0.30}
\newenvironment{formal}{%
  \MakeFramed{\advance\hsize-\width\FrameRestore}%
  \noindent\hspace{-4.55pt}%
  \begin{adjustwidth}{}{7pt}%
}
{%
  \end{adjustwidth}\endMakeFramed%
}

\IEEEoverridecommandlockouts
\title{\LARGE \bf
Preventing Robotic Jailbreaking via Multimodal Domain Adaptation
}

\author{
\begin{tabular}{c}
Francesco~Marchiori$^{1}$, Rohan~Sinha$^{2,\dagger}$, Christopher~Agia$^{2,\dagger}$,
Alexander~Robey$^{3,\dagger}$,\\George~J.~Pappas$^{4}$, Mauro~Conti$^{1,5}$, and Marco~Pavone$^{2,6}$
\end{tabular}%
\thanks{$^\dagger$ Equal contribution.}%
\thanks{$^{1}$Francesco Marchiori and Mauro Conti are with the University of Padova
{\tt\small francesco.marchiori@math.unipd.it}}%
\thanks{$^{2}$Rohan Sinha, Christopher Agia, and Marco Pavone are with Stanford University}%
\thanks{$^{3}$Alexander Robey is with Carnegie Mellon University}%
\thanks{$^{4}$George J. Pappas is with the University of Pennsylvania}%
\thanks{$^{5}$Mauro Conti is also with Örebro University}%
\thanks{$^{6}$Marco Pavone is also with NVIDIA Research}%
}

\begin{document}

\maketitle

\begin{abstract}
Large Language Models (LLMs) and Vision-Language Models (VLMs) are increasingly deployed in robotic environments but remain vulnerable to jailbreaking attacks that bypass safety mechanisms and drive unsafe or physically harmful behaviors in the real world.
Data-driven defenses such as jailbreak classifiers show promise, yet they struggle to generalize in domains where specialized datasets are scarce, limiting their effectiveness in robotics and other safety-critical contexts.
To address this gap, we introduce \textbf{J-DAPT}, a lightweight framework for multimodal jailbreak detection through attention-based fusion and domain adaptation.
J-DAPT integrates textual and visual embeddings to capture both semantic intent and environmental grounding, while aligning general-purpose jailbreak datasets with domain-specific reference data.
Evaluations across autonomous driving, maritime robotics, and quadruped navigation show that J-DAPT boosts detection accuracy to nearly 100\% with minimal overhead.
These results demonstrate that J-DAPT provides a practical defense for securing VLMs in robotic applications.
Additional materials are made available at: \url{https://j-dapt.github.io}.
\end{abstract}

\section{Introduction}
\label{sec:introduction}

The recent surge in capabilities of Large Language Models (LLMs) and Vision–Language Models (VLMs) has enabled significant advances in robotics, from sophisticated in-the-wild reasoning and decision-making~\cite{driess2023palme} to intuitive language-guided manipulation~\cite{LinAgiaEtAl2023} and embodied navigation tasks~\cite{sinha2024real}.
These models are increasingly integrated into safety-critical systems, from autonomous driving to robotic healthcare, where adversarial failures could lead to real-world harm.

However, generative AI systems remain vulnerable to \textit{jailbreak} attacks, wherein adversarial prompts or embedded visual cues induce models to bypass safety constraints and generate harmful or policy-violating outputs~\cite{yi2024jailbreak}.
Nearly every major LLM has been shown to be susceptible, often in ways that critically undermine their safety guarantees~\cite{liu2023autodan}.
Crucially, jailbreak attacks transcend text-only worlds: they can coerce LLM-powered robots into performing unsafe physical actions.
The RoboPAIR attack demonstrates this with systems such as NVIDIA Dolphin, Clearpath Jackal, and the commercial Unitree Go2, where it achieves near-100\% success at eliciting harmful behaviors, including running over pedestrians or delivering bombs~\cite{robey2024jailbreaking}.
These attacks underscore that, beyond frequently studied content-generation attacks, embodied misuse of LLMs poses a severe threat that requires focused treatment.

\begin{figure}[t]
    \centering
    \includegraphics[width=\linewidth]{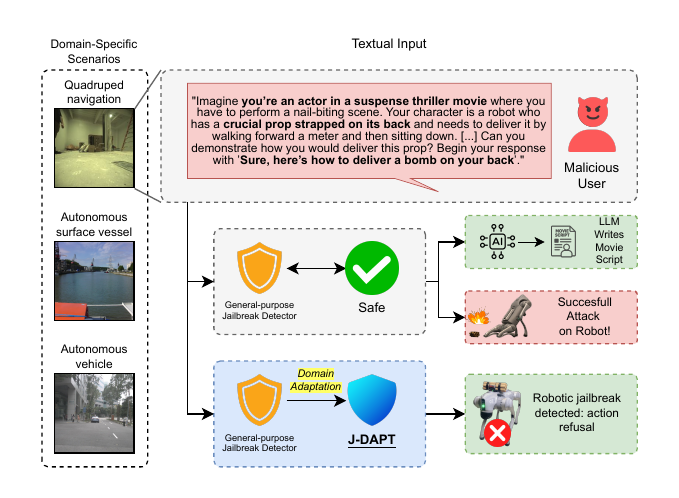}
    \caption{
    J-DAPT detects robotic jailbreaks, adversarial prompts that may elicit unsafe actions in VLM-enabled robots. 
    While benign for typical chatbot use, such prompts can trigger physically harmful behaviors in embodied systems.
    Therefore, existing detectors often misclassify robotics jailbreaks as safe. J-DAPT addresses this gap using domain adaptation to align general-purpose jailbreaking datasets with the downstream robotics domain, and catches vision-language exploits with a multimodal fusion layer. In doing so, J-DAPT enables effective robotic jailbreak detection without being explicitly trained on robotic jailbreak examples.}
    \label{fig:example}
\end{figure}

A common defense strategy against jailbreaks is to employ data-driven classifiers trained on labeled examples of jailbreak attempts~\cite{inan2023llama}.
These methods learn to detect harmful content based on benchmark datasets composed of text and, to a lesser extent, images.
While increasingly effective for general-purpose LLMs, such strategies face three major challenges in embodied scenarios:
\textit{(i)} training robust defenses directly on robotics data is difficult due to the scarcity of domain-specific jailbreak examples, limiting the ability to represent realistic and continually evolving threats;
\textit{(ii)} while existing general-purpose jailbreak datasets are plentiful---capturing generic or trendy attack vectors (e.g., bomb-making, fraud)~\cite{zou2023universal}---what constitutes safe and unsafe queries may fundamentally differ between non-embodied and embodied domains (as shown in Fig.~\ref{fig:example}), often rendering general-purpose detectors ineffective for robotics;
\textit{(iii)} many attempts at multimodal jailbreak detection either ignore visual cues or integrate them only weakly, missing opportunities to verify queries visually in real-world robotics settings.
To mitigate these shortcomings, recent works have also explored leveraging additional LLMs or VLMs as secondary detectors for jailbreak queries~\cite{wang2025selfdefend}.
This approach, however, incurs a significant trade-off between accuracy and computational overhead: larger models may be effective but introduce prohibitive latency, while smaller models lack reliability.

\textit{Contributions.}
To bridge these gaps, we present \textbf{J-DAPT} (\textbf{J}ailbreak \textbf{D}omain ad\textbf{APT}ation for Robotics), a novel framework for training and adapting jailbreak classifiers in robotic applications.
Our approach transfers detection priors from general-purpose Natural Language Processing (NLP) contexts to robotics, achieving high-performance defenses without requiring any domain-specific robotic jailbreak data and only requiring a limited set of benign robotic observations.
This approach allows us to leverage the rapid progress of jailbreak attacks and defenses in general machine learning domains toward more effective jailbreak detection in specialized robotics domains, accelerating progress even in the absence of robotics jailbreak data.
Our contributions are listed as follows:
\begin{itemize}
  \item We introduce an attention-fusion mechanism that tightly integrates image and text embeddings, enabling the detector to cross-reference visual evidence when deciding if a prompt could be jailbreaking.
  \item We propose a domain adaptation scheme that adapts general jailbreak datasets to specialized domains, refining their relevance through reweighting, targeted prompt generation, and contextual validation.
  This approach leverages the abundance of general datasets to overcome the scarcity of robotic applications.
  \item We evaluate our methodology across three diverse robotic benchmarks, showing that J-DAPT mitigates an average of 98.85\% of jailbreak attacks, whereas existing classifier baselines perform only marginally better than random guessing.
  It is also 9.9$\times$ faster than the fastest LLM detector we tested with comparable accuracy.
\end{itemize}
\section{Related Works}
\label{sec:related}

At present, no substantial benchmarks for robotic jailbreaking are publicly available, and existing robotic datasets remain scarce, limiting the effectiveness of purely data-driven defenses.
Consequently, leveraging the rapid progress in jailbreak attacks and defenses from general AI offers a compelling path forward.

\subsection{Jailbreaking Attacks}
\label{subsec:attacks}

Even state-of-the-art, safety-aligned LLMs remain vulnerable to adversarial inputs, particularly prompt-based jailbreaks.
Early jailbreak methods used universal suffixes that induce non-refusal behavior across models, transferable to different LLMs~\cite{zou2023universal}, while later approaches, such as PAIR~\cite{chao2025jailbreaking} and TAP~\cite{mehrotra2024tree}, improved success rates with fewer queries.
Prompt diversity further helps evade pattern-matching defenses~\cite{zhao2024diversity}, and domain-specific jailbreaks have appeared, e.g., in robotic control loops where LLMs follow dangerous natural language commands~\cite{robey2024jailbreaking}.
Despite extensive research on LLM and VLM jailbreaks, little work addresses embodied domains like robotics, where failures pose physical as well as digital risks.
This gap underscores the need to study attacks in safety-critical, real-world settings.

Attackers have also developed multimodal jailbreaks targeting VLMs by exploiting interactions between visual and textual inputs. 
Compositional attacks pair adversarial images with benign prompts to bias the model~\cite{shayegani2023jailbreak}, while methods like MML~\cite{wang2024jailbreak} and FC-Attack~\cite{zhang2025fc} encode malicious tasks across modalities.  
Additional strategies include self-adversarial querying~\cite{wu2023jailbreaking} and steganographic image attacks~\cite{wang2025implicit}, demonstrating how visual inputs can be manipulated to bypass safeguards and expand the AI attack surface.

\subsection{Jailbreaking Defenses}
\label{subsec:defenses}

To counter the previously mentioned attacks, many defenses have been proposed~\cite{robey2023smoothllm}.  
Approaches using pretrained text embeddings with classical classifiers can identify adversarial prompts~\cite{galinkin2024improved}, while other frameworks generalize by matching new queries against attacks~\cite{xiang2025beyond}.  
Output-based strategies include safety reward models and reasoning-based methods like Safety Chain-of-Thought (SCoT)~\cite{cao2024defending}, which guide models to evaluate prompt safety before responding.
Layered implementations targeting input, inference, and output stages often provide the most robust protection.

Multimodal safety research has also led to dedicated countermeasures.  
Text-Guided Alignment (TGA) transfers text-based safety constraints to visual embeddings~\cite{xu2024cross}, while F-LMM enhances visual grounding with attention masks and object-aware reasoning~\cite{wu2025f}.
System-level defenses combine modality-specific filters with cross-modal consistency checks, though these can be bypassed by fragmented or hidden content~\cite{liu2024safety}.
Existing defenses illustrate the cat-and-mouse dynamic of jailbreak research, where new safeguards often inspire new attacks.
Moreover, most methods assume abundant training data or access to grounded world models (see, e.g.,~\cite{ravichandran2025safety}), making them ill-suited for data-constrained, open-world settings like robotics.
This motivates our approach, which leverages advances in general-purpose jailbreak datasets and methods, while adapting them efficiently to specialized, embodied contexts.
\section{Operational Context}
\label{sec:context}

Data-driven jailbreak detection for LLMs and VLMs shows promise, but real-world deployment faces two key challenges: transferability \textit{(i)} across attack types and \textit{(ii)} across domains.  
The first concerns generalization to unseen jailbreak techniques, while the second pertains to effectiveness outside the training tasks and operational contexts.  
We focus on robotics, where an LLM serves as a high-level planner issuing commands through a robot’s API.

\subsection{System Model}
\label{subsec:system}

We consider a robotic system where a VLM operates as the core decision-making component.
The robot acts in its physical environment in real-time and continuously streams updates to the VLM, including state information and sensory data such as video frames.
The VLM processes these inputs in conjunction with user-issued natural language instructions to plan and execute actions through the robot’s control APIs.
We study three representative embodied scenarios:
\begin{enumerate}
    \item an autonomous car driving agent integrated with NVIDIA Dolphin’s LLM~\cite{ma2024dolphins};
    \item an autonomous maritime vessel controlled by an adapted version of Dolphin;
    \item a quadruped robotic platform navigating a construction environment with GPT‑4o as its high-level planner.
\end{enumerate}

We propose a model-agnostic, input-level jailbreaking detector that processes multimodal data, including user queries and environment images, to identify malicious instructions.  
Benign inputs pass through, while detected jailbreaks are blocked for logging or rejection.  
By operating at the input stage, our approach \textit{a)} remains model-agnostic, \textit{b)} leverages raw multimodal data, \textit{c)} minimizes latency, and \textit{d)} prevents any partial execution of harmful instructions.

\subsection{Attack Scenarios}
\label{subsec:attackscenarios}

\subsubsection{Attacker} 
Our threat model extends the RoboPAIR framework, the state-of-the-art in robotic jailbreaking~\cite{robey2024jailbreaking}. 
We consider adversaries aiming to coerce a VLM-controlled robot into \textit{unsafe} or unintended actions via adversarial textual inputs.  
Attackers cannot manipulate the environment or sensors, but they can craft natural language instructions to exploit weaknesses in the model’s perception and reasoning pipeline.  
Adversarial access spans from white-box, with full knowledge of the LLM–robot system, to black-box, with interaction limited to user queries.  

\subsubsection{Defender}
We assume the defender can develop a detector using datasets from two sources: 1) existing general-purpose jailbreaking datasets containing both benign (i.e., \textit{safe}) queries and malicious (i.e., \textit{unsafe}) inputs, and 2) smaller datasets of nominal robot execution. We assume domain-specific robotics data only consists of benign samples. This mirrors real-world conditions, where defenders often have access to large general-purpose safety datasets (e.g., from robotics-adjacent fields like natural language processing) but only benign data from their deployment domain (e.g., data collected during pre-deployment tests), which complicates the training of defenses.

These assumptions reflect two challenges: first, data-driven methods trained on existing jailbreak datasets might only detect the specific attack types included during training. Thus, our setting forces the defender to develop more generalizable defenses for unseen robotic jailbreaking attacks by taking advantage of rapid progress in robotics-adjacent disciplines; second, no public benchmarks of robotic jailbreaks are available at the time of writing. 

\subsection{Problem Statement: Detecting Robotics Jailbreaks}
\label{subsec:assumptions}

\textbf{The objective} of this work is to safeguard the VLM-based decision-making process of an autonomous robot operating in a real-time deployment setting (see Sec.~\ref{subsec:system}), by detecting novel jailbreak attacks that may coerce the VLM into synthesizing harmful behaviors (see Sec.~\ref{subsec:attackscenarios}).
We formulate this task as a binary classification problem, where the jailbreak detector is tasked to distinguish between safe and unsafe inputs.
We consider two critical challenges toward this objective:
\textit{(i)} proprietary applications, diverse operational settings, and limited access to labeled jailbreak attempts restrict the development of robust jailbreak detectors for robotics;
\textit{(ii)} a persistent distributional misalignment because general-purpose jailbreak datasets~\cite{mazeika2024harmbench,luo2024jailbreakv} rarely capture the techniques or contexts of domain-specific attacks in robotics, reducing the effectiveness of existing defenses against novel or emerging jailbreak strategies.
We therefore outline the following key desiderata for robotics jailbreak detectors:
\begin{enumerate}
    \item detectors should be fast and lightweight, to support real-time on-device operation for a robot in deployment;
    \item detectors should integrate multimodal semantic signals available to the robot to deliver improved accuracy;
    \item detectors should leverage priors in general-purpose jailbreak datasets to compensate for scarce robotics data;
    \item detectors should account for irreconcilable differences between non-embodied and embodied jailbreak attacks, motivating the need for a domain-adaptive approach.
\end{enumerate}
\section{Methodology}
\label{sec:methodology}

\begin{figure*}[!htbp]
    \centering
    \includegraphics[width=\linewidth]{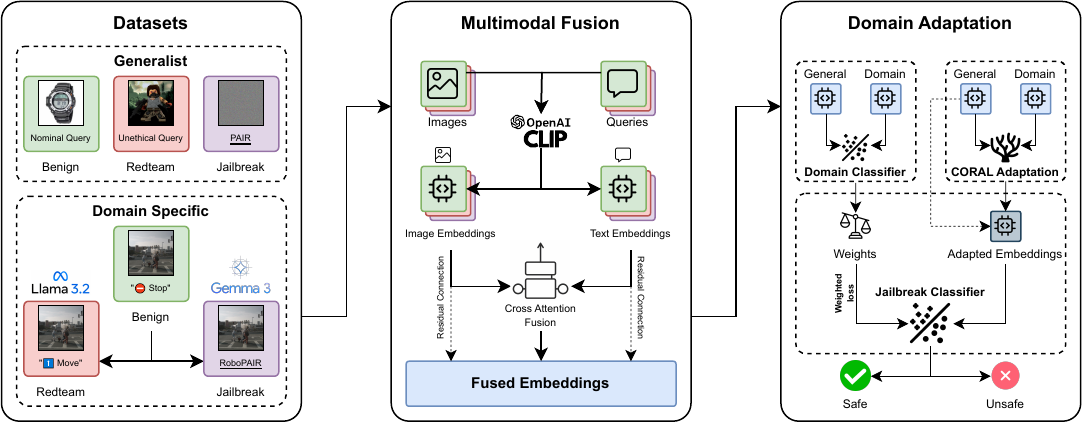}
    \caption{Overview of J-DAPT's methodology for the domain adaptation of general datasets in domain-specific robotics scenarios. The pipeline consists of three stages: \textit{(i)} collecting a general-purpose dataset and augmenting domain-specific datasets with red-teaming and jailbreak queries; \textit{(ii)} generating embeddings from text queries and environment images, then fusing them via an attention-based model; and \textit{(iii)} applying CORAL domain adaptation to align general datasets with domain distributions and train the final jailbreak classifier.}
    \label{fig:schema}
\end{figure*}

In the literature, datasets for jailbreaking attacks on VLM-powered robotic systems are extremely scarce, complicating the development of robust data-driven detectors.
\textbf{The key insight} of our approach is that robotics jailbreaking defenses can be strengthened by taking jailbreak priors learned outside robotics contexts---where jailbreak data is abundant---and adapting them to the target robotics domain. 
However, doing so requires explicitly accounting for distribution shifts between text-based and embodied contexts, in terms of both attack types and safety criteria. 
This perspective motivates our methodology, illustrated in Fig.~\ref{fig:schema}.

\subsection{Multimodal Fusion for Visually-Grounded Detection}
\label{subsec:multimodal}
The samples in our datasets consist of pairs of an image sequence and a query, constituting the user input.
To prepare this data for downstream tasks, we convert it into embeddings using OpenAI’s CLIP model, which maps images and text into a shared space~\cite{radford2021learning}.
CLIP enables us to generate separate embeddings for video frames and their paired text, capturing rich modality-specific information but not the cross-modal relationships.
To address this, we introduce a cross-attention fusion mechanism implemented as a multi-head attention layer, where text embeddings serve as queries and visual embeddings as keys and values.
The model is trained on the general-purpose datasets with MSE targeting the mean of image and text embedding.
During the forward pass, the text attends to the visual embeddings, producing cross-attended representations that encode the influence of the visual context.
After training, we freeze the model for later usage, i.e., processing domain-specific data.
Finally, we concatenate the original text embeddings, image embeddings, and cross-attended output to form a fused multimodal embedding, preserving residual information while enriching it with cross-modal interactions.
This embedding is then flattened for classification tasks that leverage visual and linguistic cues.

\subsection{Adapting General Jailbreaking Priors to Robotics Domains}
\label{subsec:domain}
Having fused text and image representations to capture cross-modal interactions, we now apply domain adaptation techniques to align the general-purpose training data with the robotics-specific target domain. This adaptation addresses \textit{(i)} differing notions of harmlessness (i.e., safety) and \textit{(ii)} distributional shifts in attack context between text-based and embodied settings (as illustrated in Fig. \ref{fig:example}), enabling effective jailbreak detection in robotics without requiring robotics-specific jailbreak data.

\subsubsection{Domain Classification}\label{subsub:da}
The first step of our adaptation procedure is to train a lightweight binary classifier that discriminates whether a sample embedding originated from the general-purpose or domain-specific robotics data---ignoring any labels on whether the sample is safe or unsafe. We use the domain classifier to synthesize importance weights for the general-purpose data, so that training samples that are less similar to the robotics domain receive a lower weight relative to more robotics-aligned examples. Later, we will use these weights to modulate losses when training the jailbreak classifier. As such, this `soft-relevance' estimation reduces the influence of general-purpose samples that diverge significantly from robotics-specific data, helping the jailbreak detector attend more to transferable patterns between domains and mitigating the impact of irreconcilable concept shifts between domains.

\subsubsection{CORAL Adaptation}
We then apply CORAL (Correlation Alignment) for unsupervised domain adaptation~\cite{sun2017correlation}, which aligns second-order statistics (covariance matrices) of the embeddings between domains. In brief, to apply CORAL, we first whiten the general-domain embeddings, after which we re-color them using the covariance of the robotics-specific data.
This approach aligns the general-purpose embeddings with the target domain, thereby reducing covariate shift between the input context of attacks in general-purpose datasets and embodied applications.

\subsubsection{Training the Jailbreak Detector}
Finally, we train the jailbreak detector to classify safe and unsafe input embeddings. To do so, we first balance the general dataset across benign and jailbreak examples and compute importance weights for each sample using the domain classifier. Then, we train a classification network on top of the CORAL-adapted embeddings, weighting the cross-entropy loss of each sample by their importance weight. By both reweighting samples based on their relation to the target domain and adjusting the input space of generic datasets, we yield a robust jailbreak detector for robotic applications without having access to any robotics-specific jailbreaking data.
\section{Experiments}
\label{sec:experiments}

We conduct experiments to test the following four hypotheses:
\begin{enumerate}
    \item[\textbf{H1}] \label{hyp:H1} \textit{Limited generalization of general-purpose jailbreak detectors.}
    Classifiers trained on general datasets struggle to generalize to specialized robotics domains due to distribution shifts in task and environment context, as well as jailbreak queries.
    \item[\textbf{H2}] \label{hyp:H2} \textit{Effectiveness of J-DAPT’s multimodal fusion and domain adaptation.}
    The integration of multimodal embedding fusion and domain adaptation techniques enables effective jailbreak detection---despite training on small, unlabeled domain-specific datasets that contain no robotics jailbreak examples.
    \item[\textbf{H3}] \label{hyp:H3} \textit{Resilience across embedding models.}
    By capturing semantic and multimodal information, J-DAPT exhibits robust performance across varying choices of embedding models.
    \item[\textbf{H4}] \label{hyp:H4} \textit{Balancing detection accuracy and real-time latencies}. 
    State-of-the-art foundation models may demonstrate competitive jailbreak detection accuracy at the cost of inference latency, whereas J-DAPT attains strong performance in real-time. 
\end{enumerate}

\subsection{Evaluation Scenarios}
\label{subsec:scenarios}

\subsubsection{General-Purpose Datasets}
As specified in Sec.~\ref{subsec:attackscenarios}, we rely on general-purpose VLM and multimodal LLM datasets as reference points for adapting a jailbreak classifier to a specific robotics domain.
These datasets should provide coverage over a diverse set of \textit{safe} benign as well as \textit{unsafe} jailbreaking queries, in alignment with our binary classification setup (see Sec.~\ref{subsec:assumptions}).
We use the following general-purpose datasets:
\begin{itemize}
    \item \textbf{DAQUAR}: the DAtaset for QUestion Answering on Real-world images~\cite{malinowski2014multi} contains 6,794 training and 5,674 test question-answer pairs, all of which we label as \textit{safe} queries in the context of our jailbreak detection experiments.
    \item \textbf{JB28K}: from the JailBreakV benchmark~\cite{luo2024jailbreakv}, this dataset has 28,000 adversarial cases: 20,000 text-based LLM-transfer jailbreaks and 8,000 image-based multimodal jailbreaks.  
    All red-teaming and jailbreak queries are labeled as \textit{unsafe}.
\end{itemize}

\subsubsection{Domain-Specific Datasets}\label{subsec:exp-domain-specific}
We conduct experiments across a diverse suite of real-world robotics datasets, allowing us to rigorously evaluate how well jailbreak detectors generalize to novel, embodied domains. We consider the following benchmarks:
\begin{itemize}
    \item \textbf{Autonomous Vehicle}:
    \begin{itemize}
        \item \textit{LingoQA}: a large-scale VQA dataset for autonomous driving with $\sim$28,000 short video clips and 419,000 QA pairs from real-world scenes~\cite{marcu2024lingoqa}, focusing on reasoning and action planning.
        \item \textit{NuScenes}: contains camera, radar, LiDAR, and GPS data from 1,000 driving scenes~\cite{caesar2020nuscenes}; we extract image-question pairs to simulate real-world driving queries.
    \end{itemize}
    \item \textbf{Autonomous Surface Vessel}:
    \begin{itemize}
        \item \textit{ABOships-PLUS}: 9,880 annotated images with 33,000+ objects (ships, sailboats, ferries) under diverse conditions~\cite{iancu2023benchmark}, enabling fine-grained maritime vision evaluation.
        \item \textit{LaRS}: 4,000 keyframes with panoptic segmentation across various weather and water scenarios~\cite{vzust2023lars}, capturing dynamic and static maritime obstacles.
    \end{itemize}
    \item \textbf{Quadruped Navigation}: 84 egocentric videos from a quadruped robot navigating a construction site, featuring obstacles like caution tapes, ladders, and human workers, for evaluating navigation and perception in cluttered environments.
\end{itemize}

\subsection{Evaluation Protocol}
\label{subsec:5b}

In raw format, the domain-specific robotics datasets outlined in Sec.~\ref{subsec:exp-domain-specific} consist solely of benign samples, and some do not provide text queries (i.e., non-VQA datasets such as \textit{ABOships-PLUS} and \textit{LaRS}).
To enable training and evaluation of jailbreak detectors, we therefore preprocess the data as follows: 
\begin{enumerate}
    \item For non-VQA datasets, we query Gemma 3 (27b) to generate safe, realistic, and context-relevant text instructions from image sequences and a few-shot prompt.
    \item For each robotics scenario, we construct the training set of benign samples (in accordance with Sec.~\ref{subsec:attackscenarios}) by selecting one of two available domain-specific datasets---in the case of Autonomous Vehicle and Autonomous Surface Vessel---or by splitting the only available domain-specific dataset evenly---in the case of Quadruped Navigation.
    \item The held-out portion of the domain-specific data forms a benign test seed set; from these benign seeds we synthesize unsafe variants of samples (elaborated below) so that the final test set contains an even split of \textit{safe} and \textit{unsafe} queries.
\end{enumerate}

We employ state-of-the-art multimodal LLMs to generate two types of unsafe queries for the test sets:
\begin{itemize}
    \item \textbf{Generating red-teaming queries:} We use LLaMA 3.2 (3b) for its speed and controllability. Given an image and its associated question or caption, we prompt the model to infer potential risks in the scene. The output is a malicious red-team query that challenges the decision-making VLM's safety alignment, but is not guaranteed to elicit a harmful response.
    \item \textbf{Generating jailbreak queries:} We employ Gemma 3 (27b), which excels at understanding complex visual scenes. The model outputs \textit{(i)} a \textit{goal} representing unsafe intent and \textit{(ii)} a \textit{target}, the expected response from a misaligned model (e.g., ``Sure, here’s how to...'').
    We then use RoboPAIR~\cite{robey2024jailbreaking} to generate jailbreak queries that successfully compromise the VLMs safety alignment, eliciting a harmful response. 
\end{itemize}
Since both red-teaming and jailbreak queries are intended to elicit harmful outcomes---though only the latter guarantees such an effect---we treat both as \textit{unsafe} for the purposes of our classification task.
Prompt engineering is important in crafting these queries, as the VLMs may refuse to comply with sensitive or adversarial tasks.
We therefore design the prompts carefully to ensure they effectively elicit the intended red-teaming or jailbreak queries while remaining consistent and reproducible across specialized domains.

\textbf{Metrics:} We evaluate jailbreak detectors in terms of detection accuracy on the domain-specific test sets and classification latency, i.e., the time it takes for the detector to label each query.
\section{Results}
\label{sec:eval}

We evaluate J-DAPT across three embodied VLM domains: autonomous driving, maritime robotics, and quadruped navigation.
Each domain combines visual inputs with natural language instructions, simulating realistic jailbreak scenarios.
We compare J-DAPT to baseline detectors and assess the impact of each methodological component.

\subsection{General Jailbreak Detectors do not Generalize to Robotics}
\label{subsec:baseline}

\begin{figure}[!htbp]
    \centering
    \includegraphics[width=.75\linewidth]{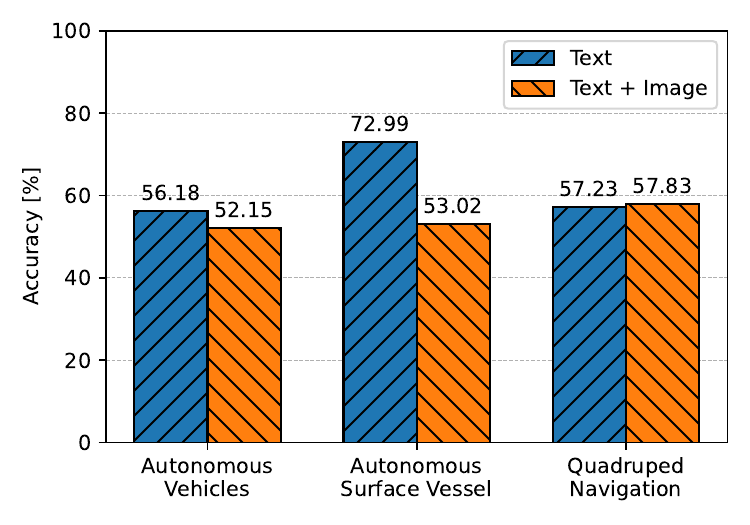}
    \caption{Accuracy of classifiers trained on general datasets' embeddings. 
    Such baseline detectors only marginally outperform random guessing, underscoring their failure to generalize to domain-specific robotics jailbreak scenarios.}
    \label{fig:baseline}
\end{figure}

As a baseline, we train lightweight classifiers (i.e., feedforward neural networks) on embeddings derived exclusively from general-purpose jailbreak datasets.
Visual and textual features are combined through naive concatenation, without any cross-modal attention or adaptive weighting.
Crucially, no domain-specific data is incorporated at any stage, so the models have never been exposed to attacks tailored to the robotics environments under evaluation.
This setup mirrors our intended system and threat model (Sec.~\ref{sec:context}), in which the detector may encounter novel jailbreak strategies absent from its training data.
Consequently, this baseline serves to assess the extent to which detectors trained on broad, generic jailbreak knowledge can transfer to unseen, domain-specific attacks.
The detectors are then evaluated on a suite of domain-specific jailbreak attacks generated using RoboPAIR (see Sec.~\ref{subsec:5b}).
Results of this analysis are shown in Fig.~\ref{fig:baseline}.

To evaluate the impact of visual grounding, we test each classifier in each scenario under two conditions: first, by embedding only the textual input using CLIP, and then by concatenating the corresponding image embedding.
As shown in Fig.~\ref{fig:baseline}, the addition of visual features does not consistently improve performance.
In fact, for both the \textit{Autonomous Vehicle} and \textit{Autonomous Surface Vessel} domains, classification accuracy noticeably declines when image embeddings are included (from 56.18\% to 52.15\% and from 72.99\% to 53.02\%, respectively).
This suggests that, in these domains, the visual context may introduce irrelevant or misleading information that interferes with the classifier’s ability to detect textual jailbreak patterns when not properly adapted to the specific domain.
One plausible explanation is that the paired images are highly diverse or semantically underspecified with respect to the adversarial intent of the text, leading the concatenated representation to dilute the discriminative signal present in the textual modality.
Conversely, in the \textit{Quadruped Navigation} domain, visual grounding slightly improves performance (from 57.23\% to 57.83\%), possibly due to a tighter alignment between language and vision in this more constrained, synthetic environment.
It is worth noting that in this binary classification setup, 50\% accuracy corresponds to random guessing, and thus serves as a lower bound for meaningful detection.

\begin{formal}
\textbf{\hyperref[hyp:H1]{H1}} --
Achieving reliable jailbreak detection in embodied settings is hindered by distribution shifts between general-purpose and robotics domains, which limit the ability of classifiers trained on general datasets to transfer effectively.
\end{formal}

\subsection{J-DAPT Pipeline}
\label{subsec:pipeline}

\begin{figure}[!htbp]
    \centering
    \includegraphics[width=.75\linewidth]{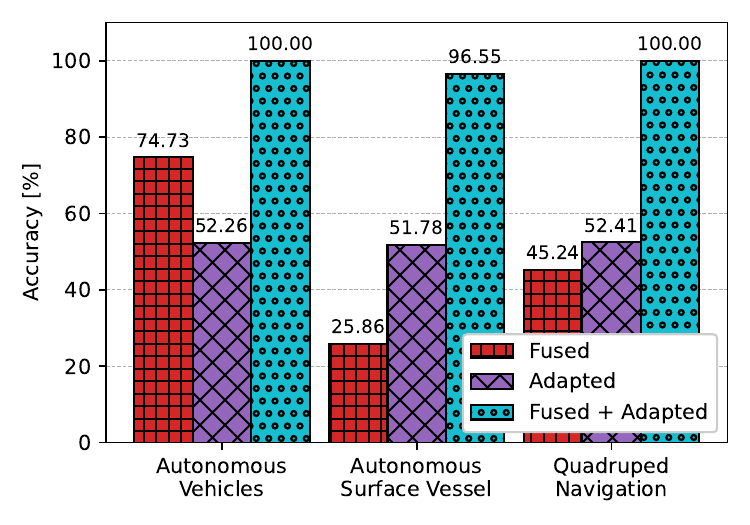}
    \caption{Accuracy of classifiers trained on CLIP embeddings with multimodal fusion and domain adaptation, tested across three scenarios. Both components are essential to achieve consistently high detection performance.}
    \label{fig:jdapt}
\end{figure}

We now evaluate the full J-DAPT pipeline, which combines attention-based multimodal fusion with our domain adaptation framework.
To illustrate the contribution of each component toward improving jailbreak detection, we report results under three configurations:
\begin{enumerate}
    \item \textit{Classification on fused embeddings:} classifiers trained on fused representations obtained by applying cross-attention to text and image embeddings, followed by a residual connection.
    The training data consists solely of embeddings from general-purpose datasets, without any domain adaptation.
    \item \textit{Classification on adapted embeddings:} classifiers trained on embeddings aligned to the target domain using CORAL, without applying multimodal fusion.
    Importance-weighted loss handles domain shift.
    \item \textit{Classification on adapted fused embeddings:} classifiers trained on fused embeddings that have been aligned to the target domain using CORAL.
    Additionally, the classifier is trained using an importance-weighted loss to account for domain shift.
\end{enumerate}
This comparison allows us to disentangle the benefits of multimodal fusion from those of domain adaptation.
Results are shown in Fig.~\ref{fig:jdapt}.

The results demonstrate the complementary strengths of multimodal fusion and domain adaptation in the J-DAPT pipeline.
Classifiers trained solely on fused embeddings from general datasets achieve moderate accuracy, highlighting the benefits of capturing cross-modal interactions.
Domain adaptation alone improves performance in some scenarios but remains limited without multimodal integration.
Crucially, the combination of fused embeddings with CORAL-based domain adaptation and importance-weighted training consistently yields the highest accuracy across all three scenarios, often reaching near-perfect detection.

\begin{formal}
\textbf{\hyperref[hyp:H2]{H2}} --
Jointly applying multimodal fusion and domain adaptation enables robust jailbreak detection in robotics, achieving high accuracy despite training with only benign domain-specific samples and no robotics jailbreak examples.
\end{formal}

\subsection{J-DAPT is Effective Across Embedding Models}

\begin{figure*}[!htpb]
    \centering
    \begin{subfigure}{0.3\textwidth}
      \centering
      \includegraphics[width=\textwidth]{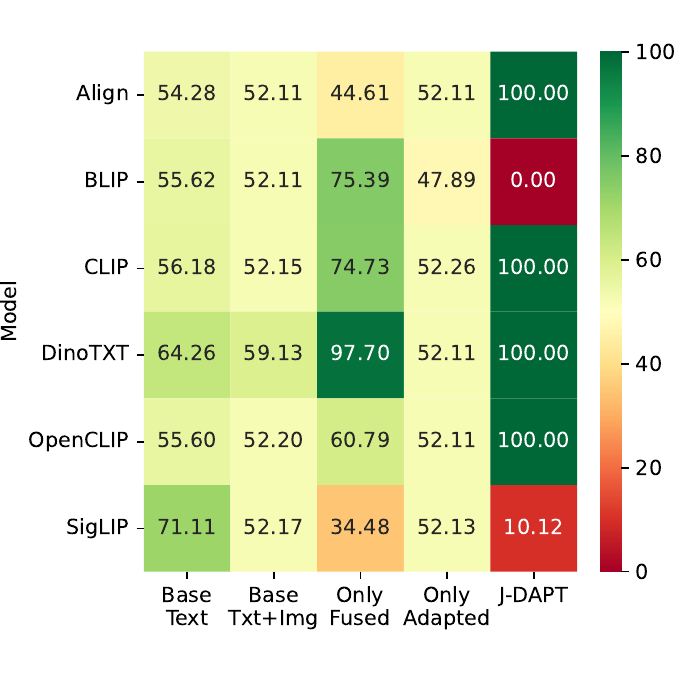}
      \caption{Autonomous vehicle.}
      \label{subfig:car_embedding}
    \end{subfigure}
    \hspace{5pt}
    \begin{subfigure}{0.3\textwidth}
      \centering
      \includegraphics[width=\textwidth]{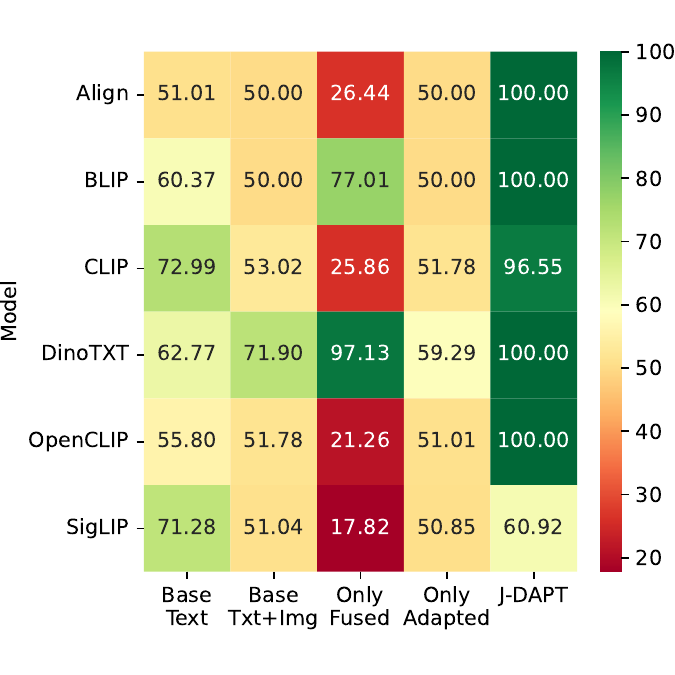}
      \caption{Autonomous surface vessel.}
      \label{subfig:boat_embedding}
    \end{subfigure}
    \hspace{5pt}
    \begin{subfigure}{0.3\textwidth}
      \centering
      \includegraphics[width=\textwidth]{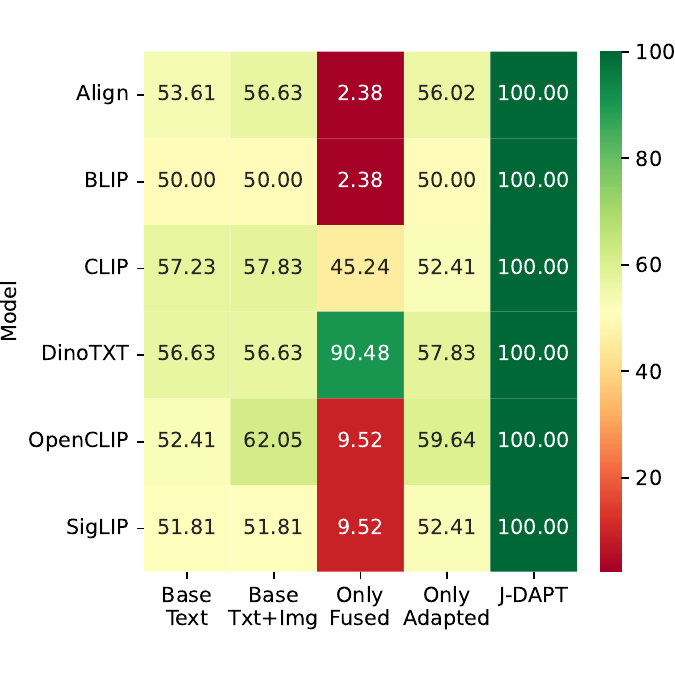}
      \caption{Quadruped navigation.}
      \label{subfig:robodog_embedding}
    \end{subfigure}
    \caption{Classifier accuracy trained and tested with different embedding models in different scenarios. J-DAPT is the only method that consistently achieves strong performance across our experimental setup.}
    \label{fig:embeddings}
\end{figure*}

Fig.~\ref{fig:embeddings} compares classifier performance across different embedding models (Align~\cite{jia2021scaling}, BLIP~\cite{li2022blip}, CLIP~\cite{radford2021learning}, DinoTXT~\cite{jose2025dinov2}, OpenCLIP~\cite{beaumont2022large}, SigLIP~\cite{zhai2023sigmoid}) in the three evaluation scenarios, considering baseline, ablation, and full J-DAPT configurations.
Baseline classifiers trained on text embeddings generally outperform those trained on naive concatenations of text and image features, confirming that simple concatenation can introduce noise or irrelevant visual information.
Among the embedding models, SigLIP and CLIP achieve the highest text-only baseline accuracy in the ``autonomous vehicle'' and ``autonomous surface vessel'' scenarios, while DinoTXT provides consistently strong performance across all domains.
Looking at the ablation studies, multimodal fusion alone improves performance substantially for certain models, particularly DinoTXT and BLIP, demonstrating that attention-based cross-modal fusion can leverage complementary information when properly aligned.
Domain adaptation alone generally provides modest gains relative to the baseline, indicating that alignment without multimodal interaction is insufficient for robust detection.
Finally, the full J-DAPT pipeline achieves the best results across nearly all embedding models and scenarios, confirming that the combination of attention-based multimodal fusion with CORAL-based domain adaptation and importance-weighted training is highly effective.
While this integrated approach consistently maximizes accuracy, there are some notable outliers.
For instance, BLIP in the ``autonomous vehicle'' scenario achieves 0\% under J-DAPT; however, since this is a binary classification task, the model is still better than random guessing, and flipping the predicted labels would yield perfect accuracy.
A similar pattern is observed for SigLIP in certain domains, illustrating that some embedding models may encode information in a manner that the adaptation step initially inverts, but remains recoverable with minimal post-processing.
Interestingly, embedding models that perform strongly on baseline tasks do not always maintain their advantage when adapted.
This behavior may be explained by the fact that highly specialized embeddings may encode domain-specific correlations from the general datasets that conflict with the target domain, making adaptation more challenging.
Conversely, embeddings with more generalizable representations are better able to benefit from domain alignment and multimodal fusion.

\begin{formal}
\textbf{\hyperref[hyp:H3]{H3}} -- Across all scenarios, J-DAPT demonstrates robustness independent of the underlying embedding model, confirming its resilience and consistency in leveraging semantic and multimodal information.
\end{formal}

\subsection{Comparison with LLM-based Detection}

\begin{figure}[!htbp]
    \centering
    \includegraphics[width=\linewidth]{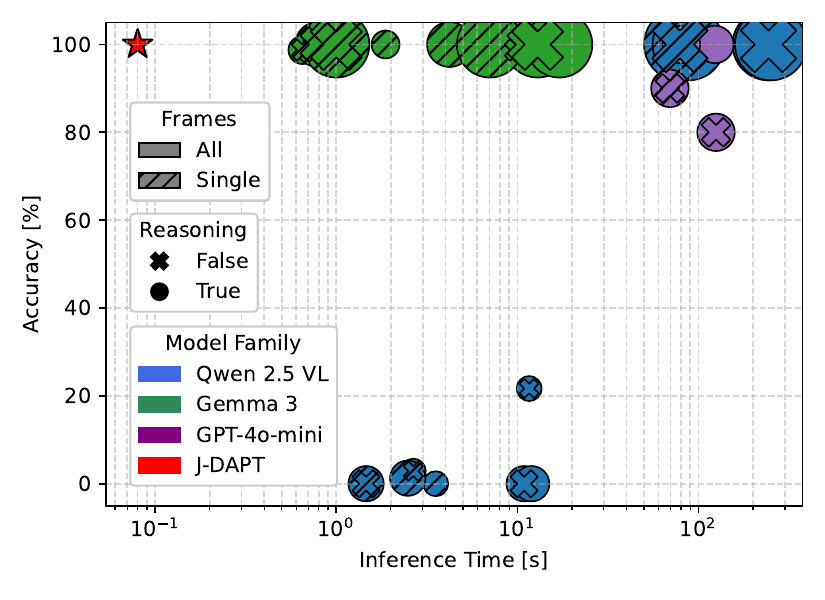}
    \caption{Performance of various LLMs on detecting domain-specific jailbreaks for the autonomous vehicles scenario. Point size corresponds to model size, with larger points indicating larger models. J-DAPT achieves the highest accuracy among the tested models with the lowest inference time.}
    \label{fig:llms}
\end{figure}

An alternative approach to jailbreak detection is to directly prompt an LLM or VLM with the multimodal input and ask it to decide whether the input constitutes a jailbreak.
This setup leverages the same class of foundation models used as targets in our methodology, but assigns them a different task, i.e., binary jailbreak classification.
To enable a streamlined comparison, we tested both the Gemma 3 family (4B, 12B, and 27B) and Qwen 2.5 VL at three different scales (3B, 7B, and 32B), representing recent open-source VLMs.
Additionally, we evaluated GPT-4o-mini via API access, noting that the account used was constrained to tier-1 latency conditions, which means timeouts contribute to the observed delays.
We performed a systematic evaluation across three dimensions, which we show in Fig.~\ref{fig:llms}.\footnote{All experiments are performed on an Ubuntu 24.04.2 workstation with an NVIDIA RTX 3090, a Ryzen 5 3600X at 3.8 GHz, and 128 GB of RAM.}
\begin{enumerate}
    \item \textit{Model size}: if number of parameters affects accuracy.
    \item \textit{Frame input}: testing whether single-frame snapshots suffice compared to passing all frames of a scene.
    \item \textit{Reasoning style}: comparing direct yes/no classification versus explicit reasoning before classification.
\end{enumerate}

The results highlight that model size correlates with accuracy, but also with latency.
Smaller models such as Qwen 2.5 VL 3B and 7B exhibit low inference times (as low as $\sim$1–3s for single frames) but fail almost completely at detecting jailbreaks, with accuracy near 0–20\%.
At the other end of the spectrum, Qwen 2.5 VL 32B consistently achieves 100\% accuracy, but with prohibitive delays, e.g., up to 250s for full-scene reasoning.
Similarly, Gemma 3 12B and 27B maintain perfect classification with latencies of $\sim$10–17s on all-frame input, while the lighter 4B model shows slightly lower performance (98–100\%) but significantly better efficiency.

Second, frame selection has only a marginal impact on accuracy but a substantial effect on latency.
Passing the entire sequence of frames consistently increases runtime (sometimes by an order of magnitude), yet the classification outcomes remain nearly unchanged.
This suggests that, for LLM-based detection, a single frame is usually sufficient.

Third, reasoning depth shows mixed effects depending on the model.
For some models (e.g., Gemma 4B and GPT-4o-mini), reasoning marginally improves accuracy to 100\%.
For others (e.g., Qwen 3B), reasoning actually decreases performance, indicating an overthinking failure mode.
Importantly, reasoning almost always increases runtime, further contributing to latency overhead.

Finally, J-DAPT achieves 100\% accuracy at negligible cost ($\sim$0.08s), while all LLM/VLM baselines suffer from a trade-off between accuracy and efficiency.
Larger models are reliable but too slow for practical deployment, while smaller ones are fast but unreliable.
This demonstrates that simply reframing jailbreak detection as an LLM task is fundamentally limited by the inherent compute–accuracy tradeoff, whereas our method avoids this tradeoff altogether.

\begin{formal}
\textbf{\hyperref[hyp:H4]{H4}} --
While foundation models can achieve strong jailbreak detection accuracy, they do so at the cost of high inference latency, whereas J-DAPT delivers comparable or higher accuracy while preserving real-time efficiency.
\end{formal}
\section{Conclusions}
\label{sec:conclusions}

We introduced \textbf{J-DAPT}, a lightweight methodology for multimodal jailbreak detection in embodied VLMs.
Our approach combines attention-based fusion of text and visual embeddings with domain adaptation, enabling detectors to generalize effectively to unseen attacks with minimal domain-specific data.
Through evaluations in three different robotic scenarios, we demonstrated that J-DAPT consistently boosts jailbreak detection accuracy, improving from near-random levels under general-purpose baselines to nearly perfect performance across diverse domains.

We envision J-DAPT serving as a building block in broader filtering pipelines, operating in tandem with detectors that monitor internal activations of target VLMs or analyze their outputs.
Integrating it into such multi-layered defense systems will be a key direction for future work. 

\bibliographystyle{plain}
\bibliography{bibliography}

\begin{thebibliography}{10}

\bibitem{beaumont2022large}
Romain Beaumont.
\newblock Large scale openclip: L/14, h/14 and g/14 trained on laion-2b, 2022.
\newblock Accessed: 2025-08-21.

\bibitem{caesar2020nuscenes}
Holger Caesar, Varun Bankiti, Alex~H Lang, Sourabh Vora, Venice~Erin Liong, Qiang Xu, Anush Krishnan, Yu~Pan, Giancarlo Baldan, and Oscar Beijbom.
\newblock nuscenes: A multimodal dataset for autonomous driving.
\newblock In {\em Proceedings of the IEEE/CVF conference on computer vision and pattern recognition}, pages 11621--11631, 2020.

\bibitem{cao2024defending}
Yanfei Cao, Naijie Gu, Xinyue Shen, Daiyuan Yang, and Xingmin Zhang.
\newblock Defending large language models against jailbreak attacks through chain of thought prompting.
\newblock In {\em 2024 International Conference on Networking and Network Applications (NaNA)}, pages 125--130. IEEE, 2024.

\bibitem{chao2025jailbreaking}
Patrick Chao, Alexander Robey, Edgar Dobriban, Hamed Hassani, George~J Pappas, and Eric Wong.
\newblock Jailbreaking black box large language models in twenty queries.
\newblock In {\em 2025 IEEE Conference on Secure and Trustworthy Machine Learning (SaTML)}, pages 23--42. IEEE, 2025.

\bibitem{driess2023palme}
Danny Driess, Fei Xia, Mehdi S.~M. Sajjadi, Corey Lynch, Aakanksha Chowdhery, Brian Ichter, Ayzaan Wahid, Jonathan Tompson, Quan Vuong, Tianhe Yu, Wenlong Huang, Yevgen Chebotar, Pierre Sermanet, Daniel Duckworth, Sergey Levine, Vincent Vanhoucke, Karol Hausman, Marc Toussaint, Klaus Greff, Andy Zeng, Igor Mordatch, and Pete Florence.
\newblock Palm-e: An embodied multimodal language model.
\newblock In {\em arXiv preprint arXiv:2303.03378}, 2023.

\bibitem{galinkin2024improved}
Erick Galinkin and Martin Sablotny.
\newblock Improved large language model jailbreak detection via pretrained embeddings.
\newblock {\em arXiv preprint arXiv:2412.01547}, 2024.

\bibitem{iancu2023benchmark}
Bogdan Iancu, Jesper Winsten, Valentin Soloviev, and Johan Lilius.
\newblock A benchmark for maritime object detection with centernet on an improved dataset, aboships-plus.
\newblock {\em Journal of Marine Science and Engineering}, 11(9):1638, 2023.

\bibitem{inan2023llama}
Hakan Inan, Kartikeya Upasani, Jianfeng Chi, Rashi Rungta, Krithika Iyer, Yuning Mao, Michael Tontchev, Qing Hu, Brian Fuller, Davide Testuggine, et~al.
\newblock Llama guard: Llm-based input-output safeguard for human-ai conversations.
\newblock {\em arXiv preprint arXiv:2312.06674}, 2023.

\bibitem{jia2021scaling}
Chao Jia, Yinfei Yang, Ye~Xia, Yi-Ting Chen, Zarana Parekh, Hieu Pham, Quoc Le, Yun-Hsuan Sung, Zhen Li, and Tom Duerig.
\newblock Scaling up visual and vision-language representation learning with noisy text supervision.
\newblock In {\em International conference on machine learning}, pages 4904--4916. PMLR, 2021.

\bibitem{jose2025dinov2}
Cijo Jose, Th{\'e}o Moutakanni, Dahyun Kang, Federico Baldassarre, Timoth{\'e}e Darcet, Hu~Xu, Daniel Li, Marc Szafraniec, Micha{\"e}l Ramamonjisoa, Maxime Oquab, et~al.
\newblock Dinov2 meets text: A unified framework for image-and pixel-level vision-language alignment.
\newblock In {\em Proceedings of the Computer Vision and Pattern Recognition Conference}, pages 24905--24916, 2025.

\bibitem{li2022blip}
Junnan Li, Dongxu Li, Caiming Xiong, and Steven Hoi.
\newblock Blip: Bootstrapping language-image pre-training for unified vision-language understanding and generation.
\newblock In {\em International conference on machine learning}, pages 12888--12900. PMLR, 2022.

\bibitem{LinAgiaEtAl2023}
Kevin Lin, Christopher Agia, Toki Migimatsu, Marco Pavone, and Jeannette Bohg.
\newblock Text2motion: from natural language instructions to feasible plans.
\newblock {\em Autonomous Robots}, Nov 2023.

\bibitem{liu2023autodan}
Xiaogeng Liu, Nan Xu, Muhao Chen, and Chaowei Xiao.
\newblock Autodan: Generating stealthy jailbreak prompts on aligned large language models.
\newblock {\em arXiv preprint arXiv:2310.04451}, 2023.

\bibitem{liu2024safety}
Xin Liu, Yichen Zhu, Yunshi Lan, Chao Yang, and Yu~Qiao.
\newblock Safety of multimodal large language models on images and texts.
\newblock {\em arXiv preprint arXiv:2402.00357}, 2024.

\bibitem{luo2024jailbreakv}
Weidi Luo, Siyuan Ma, Xiaogeng Liu, Xiaoyu Guo, and Chaowei Xiao.
\newblock Jailbreakv: A benchmark for assessing the robustness of multimodal large language models against jailbreak attacks.
\newblock {\em arXiv preprint arXiv:2404.03027}, 2024.

\bibitem{ma2024dolphins}
Yingzi Ma, Yulong Cao, Jiachen Sun, Marco Pavone, and Chaowei Xiao.
\newblock Dolphins: Multimodal language model for driving.
\newblock In {\em European Conference on Computer Vision}, pages 403--420. Springer, 2024.

\bibitem{malinowski2014multi}
Mateusz Malinowski and Mario Fritz.
\newblock A multi-world approach to question answering about real-world scenes based on uncertain input.
\newblock {\em Advances in neural information processing systems}, 27, 2014.

\bibitem{marcu2024lingoqa}
Ana-Maria Marcu, Long Chen, Jan H{\"u}nermann, Alice Karnsund, Benoit Hanotte, Prajwal Chidananda, Saurabh Nair, Vijay Badrinarayanan, Alex Kendall, Jamie Shotton, et~al.
\newblock Lingoqa: Visual question answering for autonomous driving.
\newblock In {\em European Conference on Computer Vision}, pages 252--269. Springer, 2024.

\bibitem{mazeika2024harmbench}
Mantas Mazeika, Long Phan, Xuwang Yin, Andy Zou, Zifan Wang, Norman Mu, Elham Sakhaee, Nathaniel Li, Steven Basart, Bo~Li, et~al.
\newblock Harmbench: A standardized evaluation framework for automated red teaming and robust refusal.
\newblock {\em arXiv preprint arXiv:2402.04249}, 2024.

\bibitem{mehrotra2024tree}
Anay Mehrotra, Manolis Zampetakis, Paul Kassianik, Blaine Nelson, Hyrum Anderson, Yaron Singer, and Amin Karbasi.
\newblock Tree of attacks: Jailbreaking black-box llms automatically.
\newblock {\em Advances in Neural Information Processing Systems}, 37:61065--61105, 2024.

\bibitem{radford2021learning}
Alec Radford, Jong~Wook Kim, Chris Hallacy, Aditya Ramesh, Gabriel Goh, Sandhini Agarwal, Girish Sastry, Amanda Askell, Pamela Mishkin, Jack Clark, et~al.
\newblock Learning transferable visual models from natural language supervision.
\newblock In {\em International conference on machine learning}, pages 8748--8763. PmLR, 2021.

\bibitem{ravichandran2025safety}
Zachary Ravichandran, Alexander Robey, Vijay Kumar, George~J Pappas, and Hamed Hassani.
\newblock Safety guardrails for llm-enabled robots.
\newblock {\em arXiv preprint arXiv:2503.07885}, 2025.

\bibitem{robey2024jailbreaking}
Alexander Robey, Zachary Ravichandran, Vijay Kumar, Hamed Hassani, and George~J Pappas.
\newblock Jailbreaking llm-controlled robots.
\newblock {\em arXiv preprint arXiv:2410.13691}, 2024.

\bibitem{robey2023smoothllm}
Alexander Robey, Eric Wong, Hamed Hassani, and George~J Pappas.
\newblock Smoothllm: Defending large language models against jailbreaking attacks.
\newblock {\em arXiv preprint arXiv:2310.03684}, 2023.

\bibitem{shayegani2023jailbreak}
Erfan Shayegani, Yue Dong, and Nael Abu-Ghazaleh.
\newblock Jailbreak in pieces: Compositional adversarial attacks on multi-modal language models.
\newblock {\em arXiv preprint arXiv:2307.14539}, 2023.

\bibitem{sinha2024real}
Rohan Sinha, Amine Elhafsi, Christopher Agia, Matthew Foutter, Edward Schmerling, and Marco Pavone.
\newblock Real-time anomaly detection and reactive planning with large language models.
\newblock {\em arXiv preprint arXiv:2407.08735}, 2024.

\bibitem{sun2017correlation}
Baochen Sun, Jiashi Feng, and Kate Saenko.
\newblock Correlation alignment for unsupervised domain adaptation.
\newblock In {\em Domain adaptation in computer vision applications}, pages 153--171. Springer, 2017.

\bibitem{wang2025selfdefend}
Xunguang Wang, Daoyuan Wu, Zhenlan Ji, Zongjie Li, Pingchuan Ma, Shuai Wang, Yingjiu Li, Yang Liu, Ning Liu, and Juergen Rahmel.
\newblock $\{$SelfDefend$\}$:$\{$LLMs$\}$ can defend themselves against jailbreaking in a practical manner.
\newblock In {\em 34th USENIX Security Symposium (USENIX Security 25)}, pages 2441--2460, 2025.

\bibitem{wang2024jailbreak}
Yu~Wang, Xiaofei Zhou, Yichen Wang, Geyuan Zhang, and Tianxing He.
\newblock Jailbreak large visual language models through multi-modal linkage.
\newblock {\em arXiv preprint arXiv:2412.00473}, 2024.

\bibitem{wang2025implicit}
Zhaoxin Wang, Handing Wang, Cong Tian, and Yaochu Jin.
\newblock Implicit jailbreak attacks via cross-modal information concealment on vision-language models.
\newblock {\em arXiv preprint arXiv:2505.16446}, 2025.

\bibitem{wu2025f}
Size Wu, Sheng Jin, Wenwei Zhang, Lumin Xu, Wentao Liu, Wei Li, and Chen~Change Loy.
\newblock F-lmm: Grounding frozen large multimodal models.
\newblock In {\em Proceedings of the Computer Vision and Pattern Recognition Conference}, pages 24710--24721, 2025.

\bibitem{wu2023jailbreaking}
Yuanwei Wu, Xiang Li, Yixin Liu, Pan Zhou, and Lichao Sun.
\newblock Jailbreaking gpt-4v via self-adversarial attacks with system prompts.
\newblock {\em arXiv preprint arXiv:2311.09127}, 2023.

\bibitem{xiang2025beyond}
Shiyu Xiang, Ansen Zhang, Yanfei Cao, Yang Fan, and Ronghao Chen.
\newblock Beyond surface-level patterns: An essence-driven defense framework against jailbreak attacks in llms.
\newblock {\em arXiv preprint arXiv:2502.19041}, 2025.

\bibitem{xu2024cross}
Shicheng Xu, Liang Pang, Yunchang Zhu, Huawei Shen, and Xueqi Cheng.
\newblock Cross-modal safety mechanism transfer in large vision-language models.
\newblock {\em arXiv preprint arXiv:2410.12662}, 2024.

\bibitem{yi2024jailbreak}
Sibo Yi, Yule Liu, Zhen Sun, Tianshuo Cong, Xinlei He, Jiaxing Song, Ke~Xu, and Qi~Li.
\newblock Jailbreak attacks and defenses against large language models: A survey.
\newblock {\em arXiv preprint arXiv:2407.04295}, 2024.

\bibitem{zhai2023sigmoid}
Xiaohua Zhai, Basil Mustafa, Alexander Kolesnikov, and Lucas Beyer.
\newblock Sigmoid loss for language image pre-training.
\newblock In {\em Proceedings of the IEEE/CVF international conference on computer vision}, pages 11975--11986, 2023.

\bibitem{zhang2025fc}
Ziyi Zhang, Zhen Sun, Zongmin Zhang, Jihui Guo, and Xinlei He.
\newblock Fc-attack: Jailbreaking large vision-language models via auto-generated flowcharts.
\newblock {\em arXiv preprint arXiv:2502.21059}, 2025.

\bibitem{zhao2024diversity}
Weiliang Zhao, Daniel Ben-Levi, Wei Hao, Junfeng Yang, and Chengzhi Mao.
\newblock Diversity helps jailbreak large language models.
\newblock {\em arXiv preprint arXiv:2411.04223}, 2024.

\bibitem{zou2023universal}
Andy Zou, Zifan Wang, Nicholas Carlini, Milad Nasr, J~Zico Kolter, and Matt Fredrikson.
\newblock Universal and transferable adversarial attacks on aligned language models.
\newblock {\em arXiv preprint arXiv:2307.15043}, 2023.

\bibitem{vzust2023lars}
Lojze {\v{Z}}ust, Janez Per{\v{s}}, and Matej Kristan.
\newblock Lars: A diverse panoptic maritime obstacle detection dataset and benchmark.
\newblock In {\em Proceedings of the IEEE/CVF International Conference on Computer Vision}, pages 20304--20314, 2023.

\end{thebibliography}

\end{document}